\documentclass[sigconf]{acmart}
\AtBeginDocument{%
  }

\setcopyright{acmlicensed}

\copyrightyear{2025}
\acmYear{2025}
\setcopyright{acmlicensed}\acmConference[CIKM '25]{Proceedings of the 34th
ACM International Conference on Information and Knowledge
Management}{November 10--14, 2025}{Seoul, Republic of Korea}
\acmBooktitle{Proceedings of the 34th ACM International Conference on
Information and Knowledge Management (CIKM '25), November 10--14, 2025,
Seoul, Republic of Korea}
\acmDOI{10.1145/3746252.3761473}
\acmISBN{979-8-4007-2040-6/2025/11}

\usepackage{enumitem}

\begin{document}


\title{AppAgent-Pro: A Proactive GUI Agent System for Multidomain Information Integration and User Assistance}
\author{Yuyang Zhao}
\authornote{Both authors contributed equally to this research.}
\email{zhaoyuyang@mail.ustc.edu.cn}
\affiliation{%
  \institution{University of Science and Technology of China}
  \city{Hefei, Anhui}
  \country{China}
}

\author{Wentao Shi}
\authornotemark[1]
\email{shiwentao123@mail.ustc.edu.cn}
\affiliation{%
  \institution{University of Science and Technology of China}
  \city{Hefei, Anhui}
  \country{China}
}

\author{Fuli Feng}
\authornote{Corresponding author.}
\email{fulifeng93@gmail.com}
\affiliation{%
  \institution{University of Science and Technology of China}
  \city{Hefei, Anhui}
  \country{China}
}

\author{Xiangnan He}
\email{xiangnanhe@gmail.com}
\affiliation{%
  \institution{University of Science and Technology of China}
  \city{Hefei, Anhui}
  \country{China}
}

\renewcommand{\shortauthors}{Yuyang et al.}

\begin{abstract}
    Large language model (LLM)-based agents have demonstrated remarkable capabilities in addressing complex tasks, thereby enabling more advanced information retrieval and supporting deeper, more sophisticated human information-seeking behaviors. However, most existing agents operate in a purely reactive manner, responding passively to user instructions, which significantly constrains their effectiveness and efficiency as general-purpose platforms for information acquisition. To overcome this limitation, this paper proposes AppAgent-Pro, a proactive GUI agent system that actively integrates multi-domain information based on user instructions. This approach enables the system to proactively anticipate users’ underlying needs and conduct in-depth multi-domain information mining, thereby facilitating the acquisition of more comprehensive and intelligent information. AppAgent-Pro has the potential to fundamentally redefine information acquisition in daily life, leading to a profound impact on human society. Our code is available at \url{https://github.com/LaoKuiZe/AppAgent-Pro}. The demonstration video could be found at \href{https://www.dropbox.com/scl/fi/hvzqo5vnusg66srydzixo/AppAgent-Pro-demo-video.mp4?rlkey=o2nlfqgq6ihl125mcqg7bpgqu&st=d29vrzii&dl=0}{link}\footnote{\url{https://www.dropbox.com/scl/fi/hvzqo5vnusg66srydzixo/AppAgent-Pro-demo-video.mp4?rlkey=o2nlfqgq6ihl125mcqg7bpgqu&st=d29vrzii&dl=0}}.
\end{abstract}




\maketitle

\section{Introduction}

The rapid advancement of large language models (LLMs), exemplified by GPT-4~\cite{GPT4} and Claude~\cite{claude}, has showcased significant progress in language understanding and multi-step reasoning. These capabilities have catalyzed the emergence of autonomous LLM-based agents, such as MetaGPT~\cite{Metagpt}, AutoGPT~\cite{Autogpt}, and HuggingGPT~\cite{Hugginggpt}. These LLM-based agents demonstrate remarkable proficiency in addressing complex, multi-step tasks, which mark a paradigm shift in human-information interaction by solving complex information retrieval tasks. For instance, systems include  Gemini Deep Research~\cite{Gemini}, OpenAI Deep Research~\cite{OpenAI2025DeepResearch}, and Grok3’s DeeperSearch~\cite{xAI2025Grok3} are capable of autonomously discovering, analyzing, and integrating heterogeneous web data into coherent technical reports or structured literature reviews, thereby significantly improving the efficiency and depth of knowledge acquisition. 

\begin{figure*}
  \includegraphics[width=\textwidth]{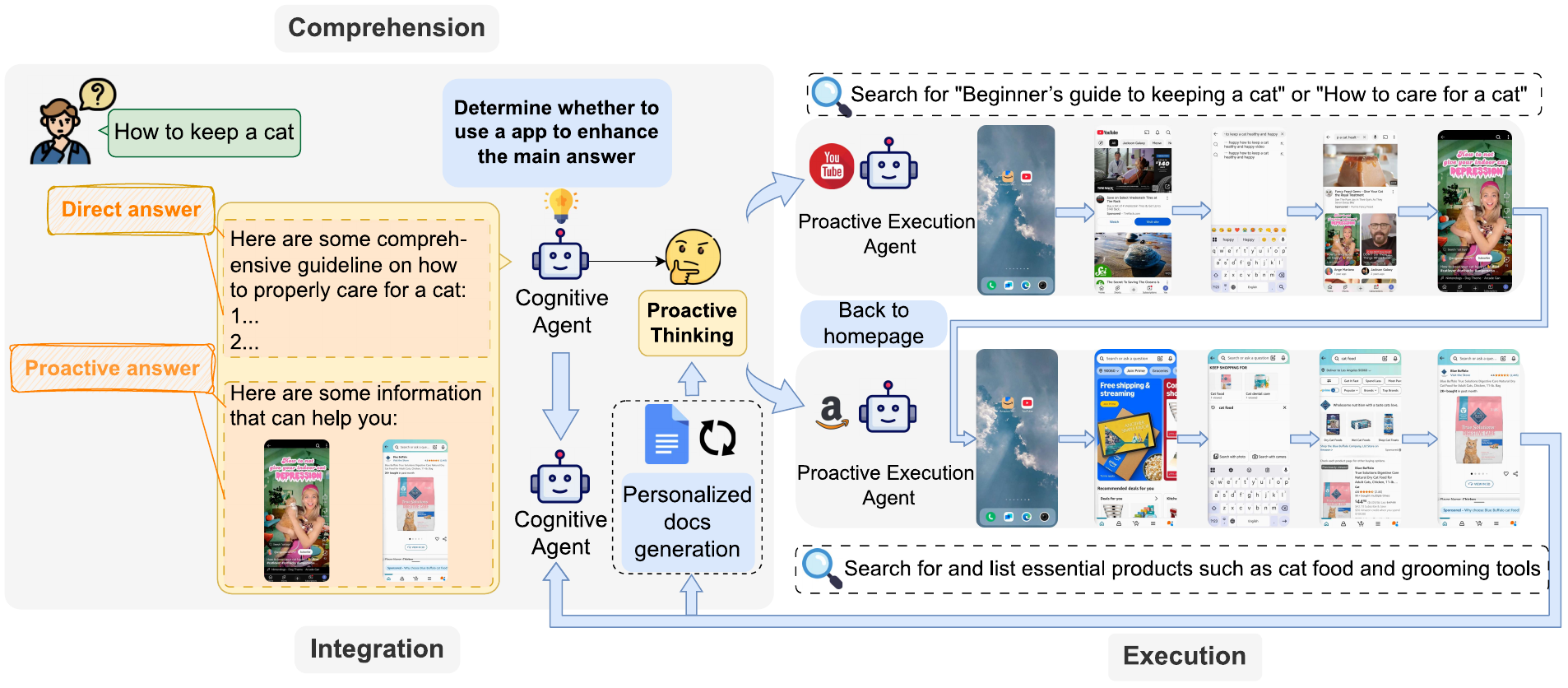}
  \caption{The architectural workflow of AppAgent-Pro, structured around a three-stage pipeline: Comprehension, Execution, and Integration. The workflow demonstrates how the system interprets and responds to open-ended user queries by proactively acquiring relevant knowledge, understanding user intent, performing appropriate actions, and integrating the results into coherent outputs.}
  \Description{The architectural workflow of AppAgent-Pro, structured around a four-stage pipeline: Learning, Comprehension, Execution, and Integration. The workflow demonstrates how the system interprets and responds to open-ended user queries by proactively acquiring relevant knowledge, understanding user intent, performing appropriate actions, and integrating the results into coherent outputs.}
  \label{fig:teaser}
  \vspace{-5pt}
\end{figure*}

However, the majority of existing LLM-based agents—particularly those operating within graphical user interface (GUI) environments—adhere to a predominantly reactive design paradigm~\cite{Mobile_agente, mobilesteward, Mobile_agentv2, Appagentx, Appagentv2}. Typically, these reactive agents execute user instructions without the capacity for autonomous planning based on environmental feedback and rely heavily on further human intervention to complete tasks. Furthermore, they only retrieve information based on explicit user inputs, failing to infer latent user intentions~\cite{Conversationalagent1, Conversationalagent2}. As a result, such reactive systems demand excessive manual involvement and impose significant cognitive burdens on users, diminishing system efficiency and degrading the overall interactive experience~\cite{Ask_before_plan, Proactive_agent}. In the era of information abundance, the reactive agents are fundamentally insufficient for realizing the potential of GUI agents as universal platforms capable of integrating and reasoning across heterogeneous information domains.


To overcome the limitations of such reactive agent paradigms, we introduce AppAgent-Pro, a proactive GUI agent system that actively integrates multi-domain information based on user instructions. AppAgent-Pro expands upon the AppAgent~\cite{Appagent} framework, enabling the system to proactively anticipate users’ underlying needs and conduct in-depth multi-domain information mining. Specifically, upon receiving a user instruction, AppAgent-Pro not only responds directly but also proactively infers the user's potential latent needs and determines whether it is necessary to retrieve supplementary information from mobile platform applications to enhance the response. When accessing app-based content, AppAgent-Pro supports the deep execution mode, enabling the agent to autonomously conduct searches, identify relevant information, and perform content analysis. Based on the initial retrieval results, the agent can refine its estimation of the user's interests and iteratively initiate follow-up searches until all pertinent information has been comprehensively aggregated. Moreover, AppAgent-Pro utilizes the personal interaction histories to adapt the task execution, ensuring more contextually relevant responses. The main contributions of the paper are summarized as follows:
\begin{itemize}[leftmargin=*]
    \item We introduce a novel proactive GUI agent system, AppAgent-Pro, to automatically anticipate users’ underlying needs and integrate multidomain information.  
    \item We propose the deep execution mode to conduct in-depth information mining.
    \item We develop a GUI agent platform to provide users with personalized and in-depth information.
\end{itemize}

The remainder of this paper is organized as follows. Section 2 presents the system architecture and operational workflow of AppAgent-Pro, detailing its proactive information integration pipeline. Section 3 outlines the demonstration plan, illustrating the system’s behavior in diverse real-world scenarios. Finally, Section 4 concludes the paper by summarizing the key contributions and discussing future directions.



\section{System Overview}
AppAgent-Pro is an agent system that enables proactive, multimodal, mobile-assistant end-to-end pipeline operations without user intervention, as shown in Figure~\ref{fig:teaser}. The entire AppAgent-Pro work process consists of three processes: Comprehension, Execution, and Integration.

\begin{figure*}
  \includegraphics[width=\textwidth]{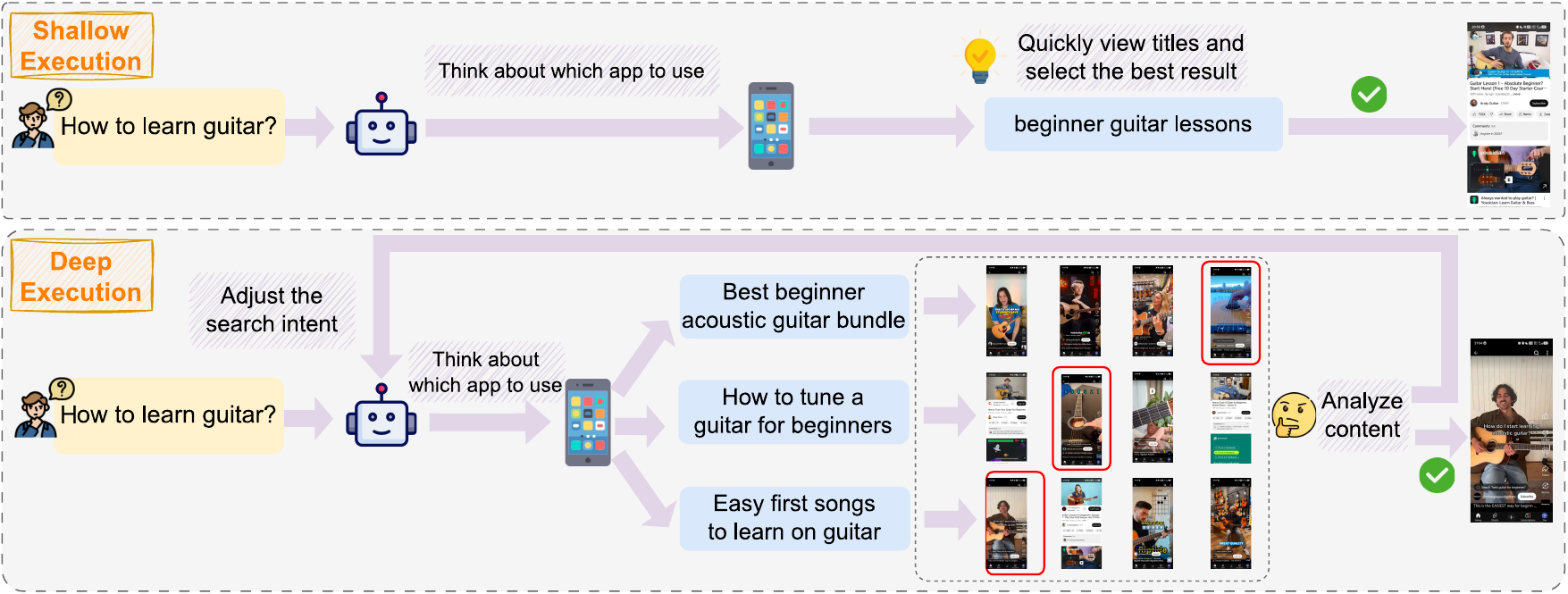}
  \caption{A comparative illustration of the shallow and deep execution modes in AppAgent-Pro. The shallow execution mode focuses on immediate interactions with limited reasoning, while the deep execution mode leverages intent-driven planning and iterative refinement to provide accurate feedback to users.}
  \Description{AppAgent-Pro follows the Learning, Comprehension, Execution, and Integration pipeline. This figure shows how AppAgent-Pro reacts to the user's open-ended query.}
  \label{fig:execution_mode}
  \vspace{-5pt}
\end{figure*}

\subsection{Comprehension}
In the comprehension stage, the system analyzes the user's instructions to identify their core requirements and latent needs, thereby determining whether proactive assistance is required. Concretely, upon receiving a user's query, such as "How to keep a cat" (as illustrated in Figure~\ref{fig:teaser}), the cognitive agent initiates a sophisticated analysis process leveraging GPT-4o. This process extends beyond basic query analysis and response, incorporating an assessment of the query’s complexity to determine whether richer and more comprehensive information can be provided by leveraging external application-based knowledge sources, such as YouTube and Amazon. The core of its proactive capability lies in predicting the user's subsequent needs and potential interactions if they were to use these applications themselves. For instance, in the cat-keeping scenario (Figure~\ref{fig:teaser}), the cognitive agent doesn't merely acknowledge the destination. Instead, it formulates specific, value-added sub-tasks for each relevant application, demonstrating an understanding of how those apps can serve the user's broader, unstated goals. The agent might determine that for YouTube, an effective sub-task is to "Search for 'Beginner's guide to keeping a cat' or 'How to care for a cat'". Simultaneously, for Amazon, it might generate sub-tasks like "Search for and list essential products such as cat food and grooming tools." These generated sub-tasks are not simplistic repetitions or direct keyword matches of the user's query. Instead, they represent a deeper level of inference, reflecting an analysis of likely user behaviors within those specific app contexts and an understanding of how each app's unique functionalities can proactively address the nuanced facets of the user's overarching request. This proactive task formulation allows the agent to gather targeted, relevant information that the user would likely seek out through manual app navigation.

\subsection{Execution}

In the execution stage, the cognitive agent autonomously interacts with applications to extract relevant information and continuously adapts based on their feedback, which is categorized into two execution modes: shallow execution and deep execution, as shown in Figure~\ref{fig:execution_mode}. In shallow execution mode, the agent selects target applications based on the user’s primary query and directly issues a search intent. It then scans the application's surface-level page titles to retrieve and rank relevant results. This mode prioritizes responsiveness and execution speed, enabling rapid delivery of preliminary results with minimal interaction overhead.

In contrast, the deep execution mode engages in a more proactive and thorough information-seeking process. The agent expands the initial user query into several sub-queries designed to anticipate potential user needs. For each sub-query, AppAgent-Pro selects relevant applications, initiates search actions, and explores deeper-level result pages. It evaluates a predefined number of result pages per sub-query and determines whether the aggregated information suffices. If the collected content is deemed insufficient, the agent dynamically generates additional sub-queries and recursively repeats the exploration process. Once adequate information is gathered, the agent compiles and returns the final results. Compared to the shallow mode, deep execution provides richer, more proactive information tailored to complex or ambiguous user intents.

\subsection{Integration}
In the integration stage, the cognitive agent integrates the multi-domain information. Upon receiving the information retrieved by the proactive execution agent, the cognitive agent synthesizes it with the initial textual response generated by the LLM and the visual content obtained during proactive exploration. The resulting output is then structured into a coherent and well-organized response, which is ultimately presented through a web-based interface.

\subsection{Personalization}
To provide personalized interactive experiences, AppAgent-Pro autonomously records and summarizes the personal interaction history during the task. Upon task completion, the detailed summary is systematically updated in an operational document, forming an accurate record of the task execution process. In subsequent interactions, AppAgent-Pro references these historical documents, enabling it to leverage accumulated knowledge to execute tasks with heightened accuracy and efficiency. This approach reduces the time required for task completion, as the agent can avoid redundant actions and instead focus on optimizing subsequent operations based on past performance. Consequently, the agent not only enhances task precision but also accelerates information retrieval for users, providing them with a faster, more reliable, and personalized experience. This continuous process of learning from past interactions allows AppAgent-Pro to evolve, leading to progressively improved efficiency and effectiveness in delivering faster, more accurate results.

\section{Demonstration Plan}
To provide an intuitive demonstration experience, we have developed an interactive web interface using Streamlit. This interface allows users to interact with AppAgent-Pro through natural language queries. The primary goal of this interactive setup is to effectively showcase AppAgent-Pro's innovative approach to proactive information retrieval and task execution within real-world mobile applications. By doing so, it emphasizes the system's significant advancements over traditional, passive text-based agents, offering a more dynamic and user-centric experience.
\begin{figure}[h]
  \centering
  \includegraphics[width=\linewidth]{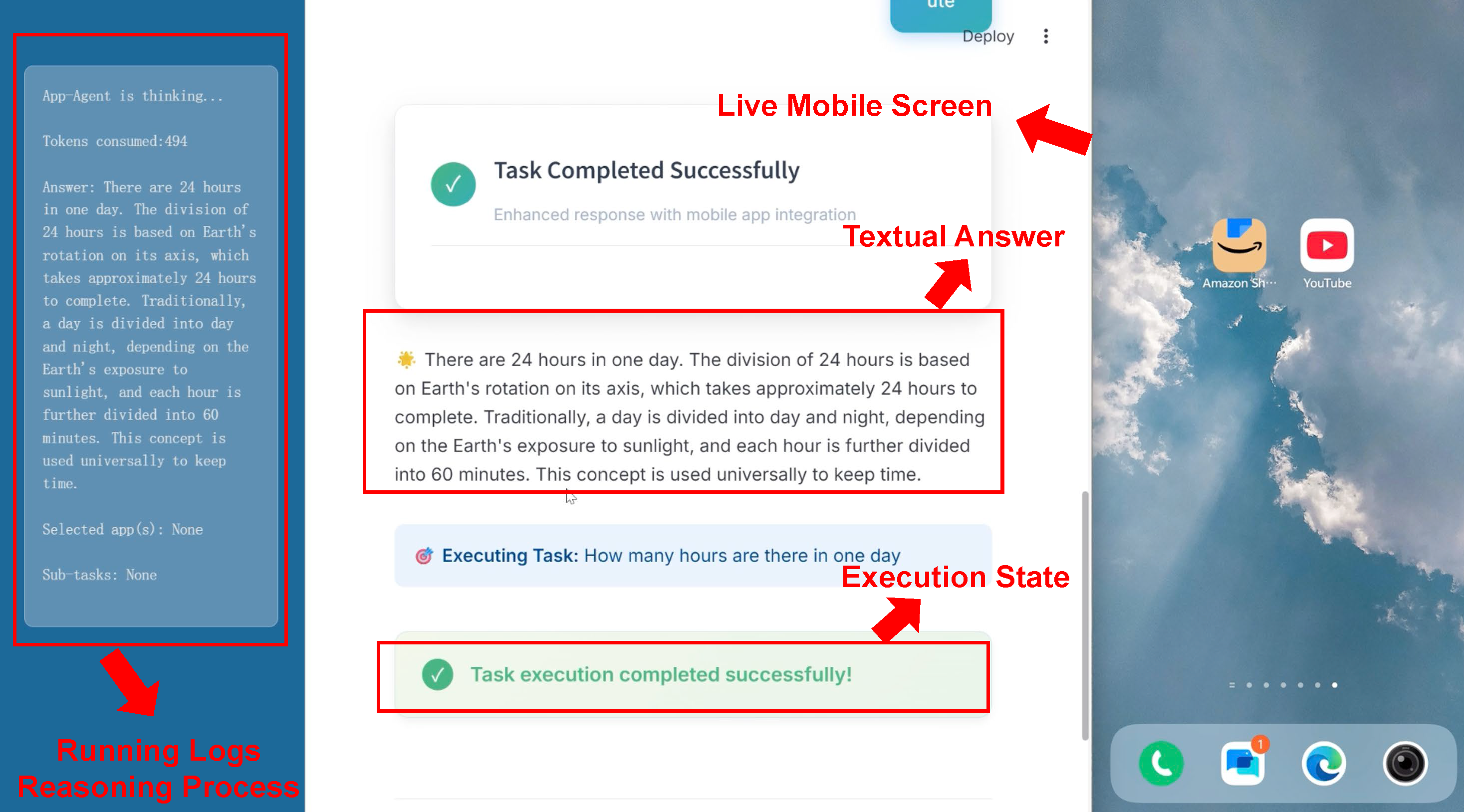}
  \caption{No external application is required for simple queries. The left column presents the reasoning process and execution log;  the middle column shows the integrated output of AppAgent-Pro, including relevant screenshots and textual responses; the right column displays the real-time mobile phone screen.}
  \Description{Mobile phone and demo web screen}
  \label{scenario1}
  \vspace{-5pt}
\end{figure}
\subsection{Scenario 1: No External App Needed}
The first scenario demonstrates AppAgent-Pro's ability to discern when internal knowledge is sufficient for the user's query. As shown in Figure~\ref{scenario1}, upon receiving a user query such as “How many hours are there in one day?”, AppAgent-Pro determines that the LLM's existing knowledge base can furnish a complete and accurate response. Consequently, it delivers the answer directly without initiating external application interactions, thereby efficiently addressing the query. This illustrates AppAgent-Pro’s capacity for judicious resource utilization, ensuring that simple knowledge-based questions are answered with minimal latency and without incurring unnecessary computational or communication overhead.



\subsection{Scenario 2: Single External App Engagement}
When the user asks, “How to upload a video on YouTube?”, AppAgent-Pro, during the understanding phase, infers that the user may require a more intuitive guide. Consequently, it activates the proactive execution agent to search for “how to upload a video” on YouTube and automatically selects the most relevant video. To further enhance usability, a screenshot of the selected YouTube video is then captured and returned as key information. The final merged response displayed on the web page will include both the textual explanation and the captured video screenshot, providing a comprehensive answer to the user's query.

\subsection{Scenario 3: Multi-App Proactive Orchestration}
\begin{figure}[!h]
  \centering
  \includegraphics[width=\linewidth]{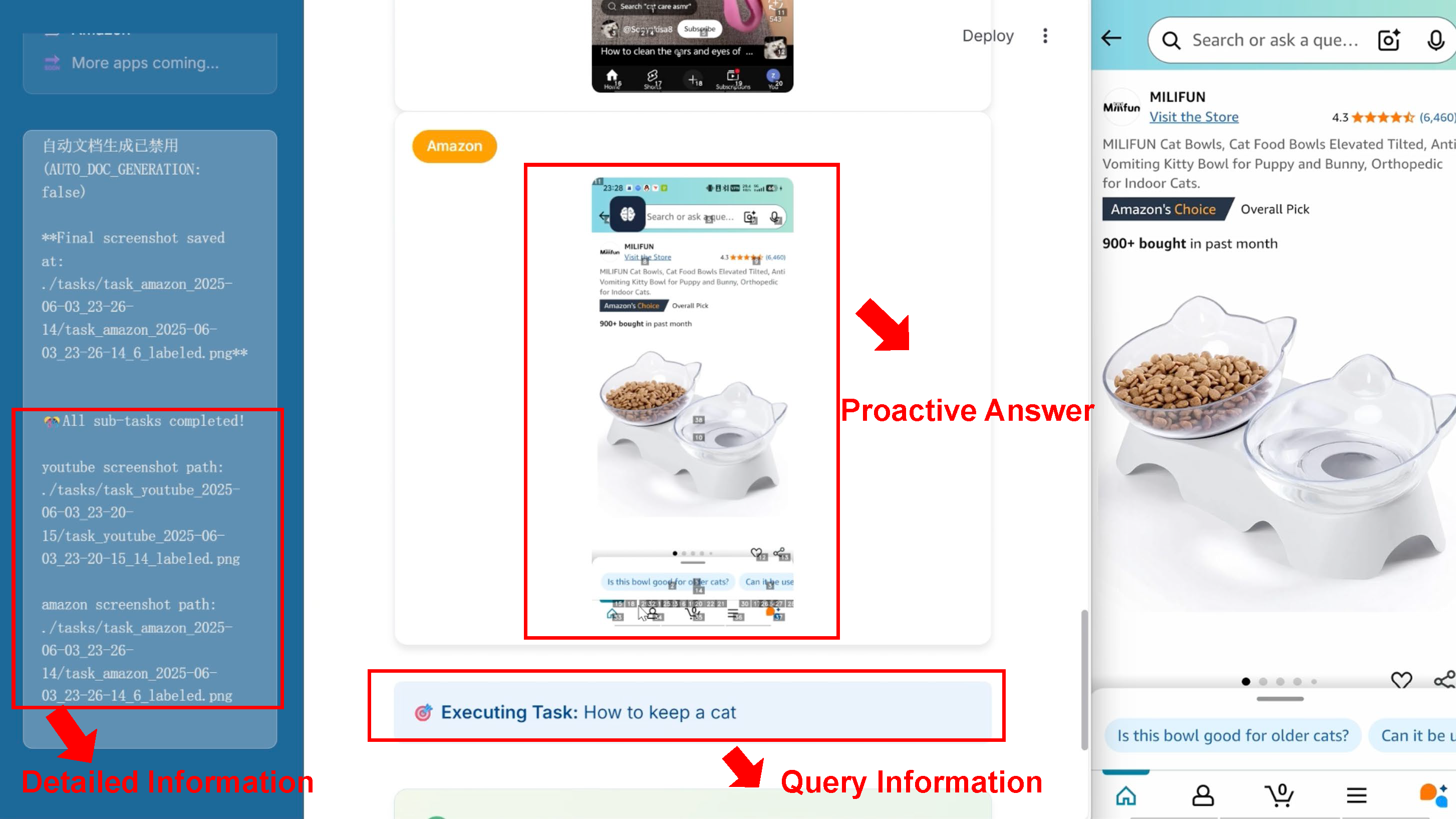}
  \caption{Proactive responses in dual-app usage. Building on the three-column layout, the left column here contains details such as file save paths and status updates; the middle column displays Amazon proactive results alongside the user query and context.}
  \Description{Mobile phone and demo web screen}
  \label{scenario3}
  \vspace{-5pt}
\end{figure}

As depicted in Figure~\ref{scenario3}, when the user asked AppAgent-Pro, “How to keep a cat?”, an open-ended query, the system deployed its most comprehensive proactive strategy. The agent identified multiple potential needs, including general cat care knowledge and basic supply information. In response, AppAgent-Pro proactively chose to leverage YouTube to search for instructional videos on cat care and keeping cats happy, while utilizing Amazon to find relevant cat supplies. Each application was assigned specific subtasks to address the user's query in a structured manner.

This example scenario effectively showcases AppAgent-Pro's advanced capabilities in decomposing complex and often implicit user requirements, orchestrating tasks across multiple disparate applications, and synthesizing a comprehensive, proactive, and multimodal response.

Although our current demonstration focuses on two representative mobile applications, the underlying architecture is highly adaptable; with minimal modifications, AppAgent-Pro can be extended to a wide range of other applications and interaction contexts.

\section{Conclusion}

In this work, we present AppAgent-Pro, a proactive GUI agent system designed to address the limitations of existing reactive LLM-based agents. By anticipating latent user intentions and autonomously integrating multi-domain information, AppAgent-Pro represents a significant advancement toward intelligent, context-aware human-computer interaction. Through its deep execution mode and adaptive retrieval mechanisms, the system enhances the efficiency, personalization, and depth of information access, paving the way for a new generation of LLM-powered GUI agents.

From a broader perspective, our findings highlight the potential of proactive agents to fundamentally reshape how users engage with complex digital ecosystems, reducing cognitive load and enabling more seamless task completion. Nevertheless, the current implementation also reveals several challenges, such as balancing proactivity with user control and maintaining robustness in dynamically evolving application environments. Future work will address these challenges by expanding the range of supported applications, incorporating more sophisticated intent inference techniques, and exploring human–AI co-adaptation strategies.


\section{GenAI Usage Disclosure}
In this work, generative AI tools were employed in a limited capacity during the preparation of this work. 
OpenAI's ChatGPT was used to assist with language refinement, including improving grammar, spelling, and readability of the manuscript text. 
Anthropic's Claude was utilized to provide suggestions for aesthetic enhancements to the demonstration webpage. 
All analyses, results, and intellectual contributions remain the sole responsibility of the authors.



\bibliographystyle{ACM-Reference-Format}
\bibliography{sample-base}

\appendix

\end{document}